\documentclass[letterpaper, 10 pt, conference]{ieeeconf}  

\IEEEoverridecommandlockouts                              

\usepackage{times}
\usepackage{amsmath}
\usepackage{amssymb}
\usepackage{xcolor}
\usepackage{pdfpages}
\usepackage{cite}
\usepackage{multirow}
\usepackage{tabularx}
\usepackage{siunitx}
\usepackage{booktabs}
\usepackage{url}
\usepackage{algorithm}
\usepackage{algorithmic}

\newcommand{\fluidsim}{FluidX3D}
\newcommand{\mapthreed}{\mathcal{M}_{3D}}
\newcommand{\occgrid}{\mathcal{O}}
\newcommand{\windreport}{\mathcal{W}}
\newcommand{\fogmap}{\mathcal{F}}

\newcommand{\ALG}{\texttt{WESPR}}
\newcommand{\BASE}{\texttt{Base}}
\newcommand{\ENV}[2][]{%
  \def\longname{}%
  \ifnum#2=6 \def\longname{\texttt{Environment-1}}\def\shortname{\texttt{E1}}\else
  \ifnum#2=3 \def\longname{\texttt{Environment-2}}\def\shortname{\texttt{E2}}\else
  \ifnum#2=4 \def\longname{\texttt{Environment-3}}\def\shortname{\texttt{E3}}\else
    \def\longname{\textcolor{red}{Unknown-Environment}}%
    \def\shortname{\textcolor{red}{Unknown-Env}}%
  \fi\fi\fi
  \if\relax\detokenize{#1}\relax
    \longname
  \else
    \shortname
  \fi
}

\title{\LARGE \bf
WESPR: Wind-adaptive Energy-Efficient Safe Perception \& Planning for Robust Flight with Quadrotors
}

\author{Khuzema Habib*$^{1}$, Pranav Deshakulkarni Manjunath*$^{1}$, Kasra Torshizi$^{1}$, Pratap Tokekar$^{1}$, Troi Williams$^{1}$%
\thanks{* Equal Contribution}%
\thanks{$^{1}$
       University of Maryland, College Park. {\tt\small \{khabib, pdeshaku, ktorsh,tokekar,troiw\}@umd.edu}.}%
}

\begin{document}

\maketitle
\thispagestyle{empty}
\pagestyle{empty}

\begin{abstract}

Local wind conditions strongly influence drone performance: headwinds increase flight time, crosswinds and wind shear hinder agility in cluttered spaces, while tailwinds reduce travel time. Although adaptive controllers can mitigate turbulence, they remain unaware of the surrounding geometry that generates it, preventing proactive avoidance. Existing methods that model how wind interacts with the environment typically rely on computationally expensive fluid dynamics simulations, limiting real-time adaptation to new environments and conditions. To bridge this gap, we present \emph{\ALG{}}, a fast framework that predicts how environmental geometry affects local wind conditions, enabling proactive path planning and control adaptation. Our lightweight pipeline integrates geometric perception and local weather data to estimate wind fields, compute cost-efficient paths, and adjust control strategies--all within 10 seconds. We validate \ALG{} on a Crazyflie drone navigating turbulent obstacle courses. Our results show a 35.6\% reduction on average in maximum trajectory deviation and a 24.6\% improvement in stability compared to a wind-agnostic adaptive controller.
\end{abstract}


\section{Introduction}\label{sec:intro}

Unmanned Aerial Vehicles (UAVs) are increasingly used in agriculture, inspection, surveillance, and delivery. Quadrotors offer agility and hovering capability but trade efficiency for maneuverability; unlike fixed-wing aircraft that passively generate lift, they must continuously expend thrust to maintain position. Turbulent winds further increase power demand and degrade stability. Larger batteries can extend flight time, but their added weight offsets endurance gains, while modern propulsion systems already operate near peak efficiency \cite{7426988}, limiting hardware improvements. Stability is also critical when transporting sensitive payloads. However, traditional planners are wind-agnostic, reacting to disturbances rather than avoiding them. Geometric controllers~\cite{lee2010geometric} and disturbance observers~\cite{Bauersfeld__2021,bauersfeld2021mpc,10313083} improve gust rejection  but remain fundamentally corrective. To address this, a planner that anticipates wind conditions could route the vehicle through calmer regions, improving both efficiency and stability.

\begin{figure}[t]
    \centering
    \includegraphics[width=0.9\linewidth]{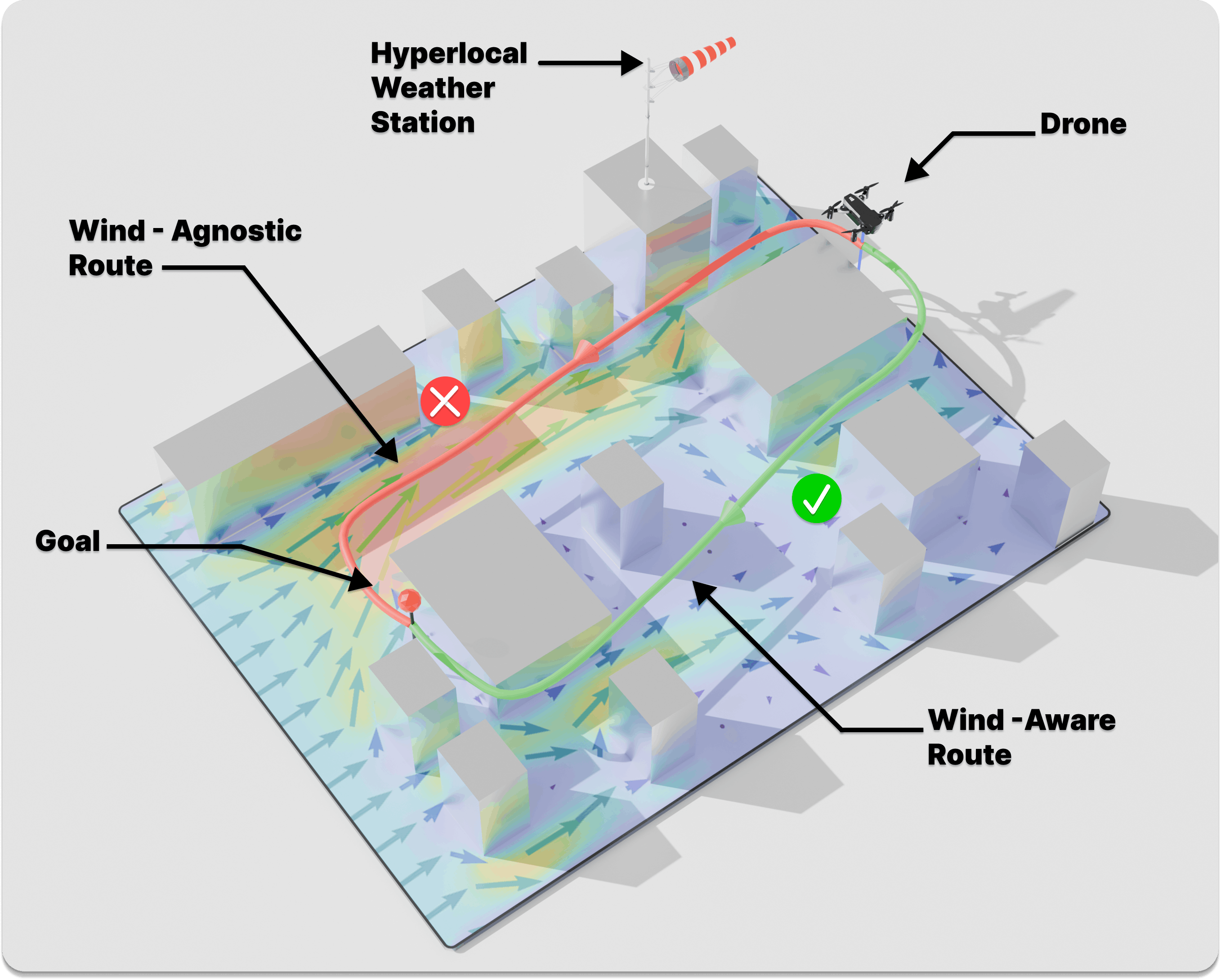}
    \caption{A UAV navigating a turbulent environment. The red path shows the shortest route but crosses stronger winds; the green path follows a wind-aware trajectory through calmer regions.}
    \label{fig:main_figure}
\end{figure}

In dense natural or urban settings, sheltered low-turbulence zones often form behind obstacles. With advancing hyper-local sensing via NOAA HRRR~\cite{james2022high}, CBAM~\cite{2024AMS...10431465D}, and wind-LiDAR~\cite{nabi2020improving}, wind-aware cost maps can be generated using Computational Fluid Dynamics (CFD) simulations. These maps penalize paths through turbulence and leverage tailwinds to reduce power consumption.  

Our research explores whether current wind field priors from a hyper-local weather station can enable energy-efficient, safer UAV path planning. Traditional CFD tools like OpenFOAM~\cite{jasak2007openfoam} are too slow for rapid deployment due to meshing and solver overhead. We instead use FluidX3D~\cite{lehmann2022fluidx3d}, a GPU-accelerated lattice Boltzmann solver that produces steady-state wind fields in seconds.  

We propose \ALG{}, an end to end, pre-flight wind-aware planning framework that integrates depth based environment reconstruction, fed into a fast CFD simulation and wind-informed path optimization (Figure~\ref{fig:main_figure}). Our paper contributes:
\begin{itemize}
    \item Rapid wind field prediction using \fluidsim\ with wind source priors.
    \item Wind-aware cost mapping based on wind velocities.
    \item Experimental validation showing reduced trajectory deviation and improved robustness under wind disturbances.
\end{itemize}

\section{Related Work}\label{sec:related}

Autonomous navigation of UAVs in outdoor environments remains challenging under wind disturbances. Small quadrotors are particularly vulnerable due to their low mass and limited airspeed, driving research in wind-aware planning, estimation, and disturbance-adaptive control.

\subsection{Wind-Aware Path Planning}
Wind-aware planning spans geometric, sampling-based, graph-search, and optimization-based methods. Geometric planners incorporate wind and vehicle constraints \cite{7963609,7152406} but assume simplified wind fields. Sampling-based approaches handle non-uniform winds from coarse global data \cite{Oettershagen2017TowardsFE}, while graph-search methods use wind-aware cost functions \cite{GU2026103605} but depend on externally provided wind maps. Optimization-based planners compute energy-efficient trajectories under wind disturbances \cite{gasche2025energy,drones7010058} but presume known flow fields. The most relevant prior work \cite{10644753} demonstrated CFD-informed Crazyflie planning, but with an offline, static wind model limiting adaptability. In contrast, \ALG{} extracts geometry from overhead perception and performs fast, physics-based wind simulation conditioned on the observed layout.

\subsection{Learning-Based Fluid Dynamics}
CNN-based models \cite{8793547} and convolutional autoencoders \cite{drones9110791} accelerate flow prediction but lack integration into hardware-validated planning pipelines. We explored PhysicsNeMo/DoMINO \cite{physicsnemo2023}, but their setup complexity and per-geometry optimization to enforce physical consistency (e.g., no-slip conditions on arbitrary STL layouts) negated inference-time gains, often requiring minutes of manual tuning per environment. We therefore employed a GPU-accelerated LBM solver that generates zero-shot, physics-verified wind fields in around 5~seconds (4~seconds on an \textit{RTX 3080}, 6~seconds on a \textit{GTX 1650}), offering reliable performance across unseen geometries.

\subsection{Environment-Aware Robotic Control}
Reactive controllers using UKF \cite{paris2020control} and EKF \cite{Stastny2019OnFB} achieve robust tracking but lack predictive, geometry-conditioned wind modeling. To our knowledge, no prior system unifies (1) geometry extraction from perception, (2) physics-based wind simulation, (3) wind-informed trajectory generation, and (4) hardware validation under controlled conditions---all of which \ALG{} integrates into a reproducible pipeline.

\subsection{Wind Estimation for UAVs}

Wind estimation research primarily targets online disturbance identification at the vehicle level. EKF-based methods using drag-augmented quadrotor models estimate local wind components from onboard sensors~\cite{7759336}. NeuroMHE integrates neural networks with Moving Horizon Estimation for adaptive parameter tuning~\cite{10313083}, while Contextual NeuroMHE extends generalization across wind regimes via Bayesian optimization~\cite{Torshizi2025ContextualNeuroMHE}. Deep reinforcement learning has also been applied for wind-adaptive tracking with hardware validation~\cite{10099019}. These approaches enhance reactive disturbance rejection by estimating local wind at the vehicle state. 

In contrast, \ALG{} predicts geometry-conditioned planar wind flow prior to execution, enabling proactive trajectory planning. Wind estimation and adaptive control methods thus complement—but do not replace—environment-conditioned flow prediction for planning.

\section{The \ALG{} Pipeline}\label{sec:method}

\begin{figure}[t]
    \centering
    \vspace{4pt}
    \includegraphics[width=0.85\linewidth]{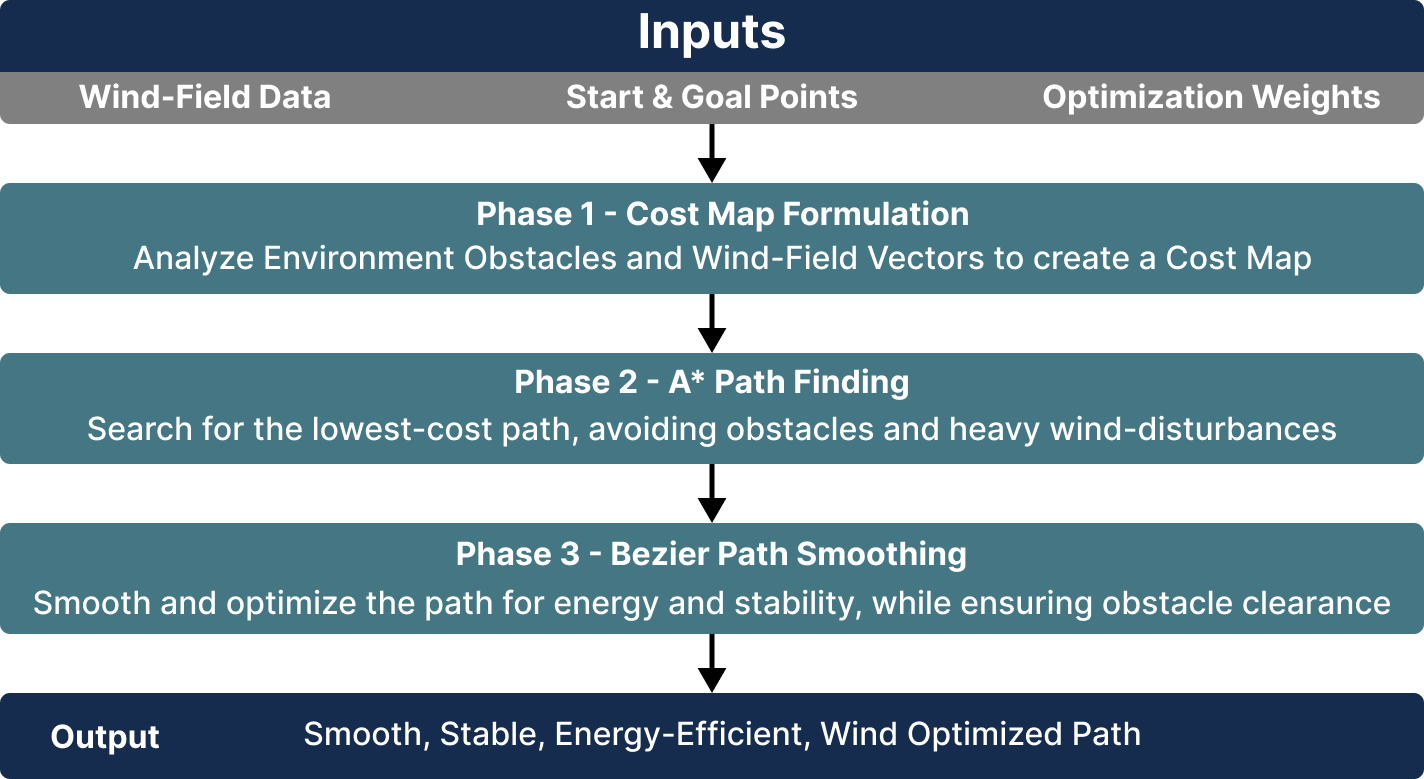}
    \caption{The \ALG{} path generation pipeline as described in Section \ref{sec:method}.}
    \label{fig:path_gen_pipeline}
\end{figure}

We now describe \ALG{}, a fast, wind field estimation and path planning pipeline (Figure \ref{fig:path_gen_pipeline}). Our pipeline is composed of 3D Mapping and Weather Modules, a Fluid Estimation Module and a Trajectory Planning Module.

\subsection{Obtaining a Local 3D Map and Wind Report}\label{sec:method:mapping_wind}

Our pipeline begins with obtaining a 3D map of the local environment $\mapthreed$. The 3D map $\mapthreed$ may be represented as a point cloud or depth map, and can be obtained from prior geometric data such as city schematics, satellite imagery, or aerial surveys \cite{wheretomap2020sasaki}. Next, we compute $\occgrid$, a 2D occupancy grid, by transforming $\mapthreed$ into a bird's eye view map, applying a depth filter for obstacles that are placed at given altitude at which the UAV would operate at $h_{min}$, and applying post-processing to remove artifacts.

At the same time, we retrieve the wind conditions ${\windreport = \{\textbf{w}_1, \ldots, \textbf{w}_n\}}$ for the local map. Each report $\textbf{w}_i$ contains the velocity of a wind source and its location in the map. We can obtain such reports from (hyper-)local weather stations or visual aids like wind socks.





\subsection{Wind Field Generation Using \fluidsim}

Given the $\occgrid$ and $\windreport$, we compute a 2D wind field $\fogmap$ using \fluidsim, a Lattice Boltzmann Method (LBM) simulator, to enable wind-aware path planning. The wind field (specifying wind speeds and directions) is defined over the discretized environment with obstacles. The simulation runs for 10,000 timesteps in about five seconds on a modern GPU and assumes a steady-state solution.


\subsubsection{Domain Construction and Boundary Conditions}


To begin computing $\fogmap$, we automatically generate a simulation domain from the discretized 2D occupancy grid $\occgrid$. Within $\occgrid$, we treat occupied cells as solid nodes and free cells as fluid nodes. Then the local winds $\windreport$ are mapped to velocity inlet boundary nodes (equilibrium node), which describe where winds enter the environment and their velocities. This representation removes the need to generate meshes, enabling procedural and automated construction of the environment.


\subsubsection{Lattice Boltzmann Formulation}




An LBM solves the discretized Boltzmann transport equation on a Cartesian lattice \cite{PhysRevE.106.015308}. The particle distribution function $f_i$ evolves via:
\begin{equation}
f_i(\mathbf{x}+\mathbf{c}_i\Delta t, t+\Delta t)
=
f_i(\mathbf{x},t)
-
\frac{\Delta t}{\tau}
\left(f_i - f_i^{eq}\right),
\end{equation}
where $\mathbf{c}_i$ are discrete lattice velocities, $\tau$ is the relaxation time, and $f_i^{eq}$ is the equilibrium distribution. Macroscopic values (density and velocity) are recovered through moment summation:
\begin{equation}
\rho = \sum_i f_i,
\qquad
\mathbf{u} = \frac{1}{\rho} \sum_i f_i \mathbf{c}_i.
\end{equation}

For 2D velocity discretization, we employ a D2Q9 lattice configuration. An LBM’s streaming-collision structure is highly parallelizable and optimized for GPU execution, enabling near real-time wind field approximation \cite{PhysRevE.106.015308}.

By bypassing the complex meshing and iterative pressure-velocity coupling required by Reynolds-Averaged Navier-Stokes or Large-Eddy Simulation solvers (e.g., $k$--$\omega$ SST or SIMPLE), LBM significantly reduces computational latency. This efficiency makes it ideal for pre-flight wind estimation within high-speed robotics planning pipelines.

\subsubsection{Physical Scaling and Unit Interpretation}

LBM operates in lattice units, where spatial and temporal quantities are non-dimensional. Physical interpretation is obtained by scaling inlet velocities by a factor of $0.5f$ to $2f$.

Let $L$ denote a reference length (e.g., obstacle width) and $U$ a reference inlet wind velocity. The Reynolds number is preserved via $Re = \frac{U L}{\nu}$, where the kinematic viscosity $\nu$ is determined by the relaxation time $\tau$.
Because the objective is relative wind velocity for path planning and not high-fidelity turbulence modeling, we operate in low-to-moderate Reynolds regime, setting $Re$ = 250 for our experimental conditions, which served as a critical simulation proxy that balances computational stability with representative flow behavior, specifically fluid-structure interactions (FSI) around obstacle geometry. This value was empirically found to resolve dominant wake structures and velocity gradients without introducing numerical instability into the LBM solver. The resulting wind-field provides sufficient fidelity for cost-map generation.



\subsection{Wind-Aware Cost Map Generation}\label{sec:method:costmapgen}

Next, we convert the 2D wind field $\fogmap$ into a cost map used for path planning (Phase 1 in Algorithm \ref{alg:hybrid-planning}). The costs penalize energetically unfavorable regions, such as strong headwinds and crosswinds, while preserving obstacle avoidance and safety margins.

\subsubsection{Projecting CFD Field onto Planning Grid} The simulated horizontal velocity components $(v_x, v_y)$, the corresponding speed magnitude $s = \|\mathbf{u}\|$, and the wall mask are interpolated onto a 2D planning grid.


The resulting grid defines the domain $\Omega$ over which path planning is performed.

\subsubsection{Obstacle Representation and Safety Buffer} The interpolated wall mask forms the initial obstacle set. A safety margin is applied by performing binary dilation with a buffer radius $b$ (specified in meters and converted to grid cells). Additionally, domain boundary cells are marked as obstacles to prevent edge traversal.

The final obstacle set is denoted by $\mathcal{O}$. All cells $(i,j) \in \mathcal{O}$ are assigned a fixed large cost $C_{ij} = c_{\text{wall}}$. This ensures that the planner maintains a clearance margin from physical structures.

\subsubsection{Goal Direction} A global unit direction vector from start to goal is computed once and is used to evaluate wind alignment across the domain.

\subsubsection{Per-Cell Flow-Aware Cost}\label{subsec: costs} When wind-aware planning is enabled, each non-obstacle cell incurs a cost derived from local wind magnitude and direction. The total cell cost combines three weighted components ${C_{ij} = w_s c_s + w_d c_d + w_a c_a}$, where $c_s = s_{ij}/\max_{\Omega} s$ penalizes high wind speed, $c_d = 1 - (\alpha_{ij}+1)/2$ penalizes headwinds ($\alpha_{ij} = \hat{\mathbf{u}}_{ij}\cdot\mathbf{g}$ ( where $\mathbf{g}=(\mathbf{p}_{\text{goal}}-\mathbf{p}_{\text{start}})/\|\mathbf{p}_{\text{goal}}-\mathbf{p}_{\text{start}}\|$ is the alignment of the start-goal), and $c_a = 1 - |\alpha_{ij}|$ penalizes crosswinds.

The cost is clamped to $[0.1, 20]$ and smoothed with a Gaussian kernel. Obstacle cells receive a fixed penalty $c_{\text{wall}}$.

The key tunable parameters are: $w_s$: speed penalty weight, $w_d$: direction (headwind) penalty weight,  $w_a$: alignment (crosswind) penalty weight,  $c_{\text{wall}}$: obstacle penalty, and $b$: wall buffer distance.

When flow-aware planning is disabled, all non-obstacle cells use a constant base cost (0.5), reducing A* to Euclidean shortest-path planning (Phase 2 in Algorithm \ref{alg:hybrid-planning}). For globally adverse winds (goal alignment $<-0.3$), the planner switches to an against-flow variant with additional edge penalties.

\begin{algorithm}[t]
\caption{\ALG{} Planning Pipeline}
\label{alg:hybrid-planning}
\begin{algorithmic}[1]
\REQUIRE Wind field $\mathbf{u}_{wind}$, Start $\mathbf{P}_0$, Goal $\mathbf{P}_n$, Weights $\lambda$
\ENSURE Optimized Bézier Curve $\mathbf{r}(s)$

\STATE \textbf{Phase 1: Discrete Cost-Map Generation}
\FOR{each free cell $(i,j)$}
    \STATE Calculate alignment $A = \frac{\mathbf{u} \cdot \vec{g}}{\|\mathbf{u}\|}$ and speed $V = \frac{\|\mathbf{u}\|}{\max \|\mathbf{u}\|}$
    \STATE $C_{i,j} \gets \text{clamp}\big(w_s V + w_d(1-\frac{A+1}{2}) + w_a(1-|A|), 0.1, 20\big)$
\ENDFOR
\STATE Set $C = \text{wall\_penalty}$ on dilated obstacle grid $\mathcal{O}$.

\STATE \textbf{Phase 2: Global Path Seeding (A*)}
\STATE Search for discrete path $\mathcal{P}_{A^*}$ using cost $C$.
\IF{Strong upstream flow ($A < -0.3$)}
    \STATE Apply $2.5\times$ velocity penalty to neighbor costs.
\ENDIF

\STATE \textbf{Phase 3: Continuous Bézier Optimization}
\STATE Initialize control points $\{\mathbf{P}_1, \dots, \mathbf{P}_{n-1}\}$ from $\mathcal{P}_{A^*}$ waypoints.
\STATE Define objective function $J = \lambda_p J_{\text{path}} + \lambda_s J_{\text{snap}} + \lambda_t J_{\text{thrust}} + \lambda_w J_{\text{wall}}$.
\STATE Solve for optimal control points $\mathbf{P}_i$ via Block Coordinate Descent:
    \begin{itemize}
        \item $J_{\text{path}}$: Minimize distance to $\mathcal{P}_{A^*}$.
        \item $J_{\text{snap}}$: Minimize fourth-order derivative (smoothness).
        \item $J_{\text{thrust}}$: Optimize for wind-aware energy using $\mathbf{u}_{wind}$.
        \item $J_{\text{wall}}$: Enforce clearance via convex hull property.
    \end{itemize}

\RETURN $\mathbf{r}(s) = \sum_{i=0}^{n} B_{i,n}(s)\mathbf{P}_i$
\end{algorithmic}
\end{algorithm}

\subsection{Bezier Trajectory Formulation and Optimization}

The discrete waypoint sequence produced by the A* planner is refined into a dynamically smooth trajectory using a parametric Bézier curve formulation, following the approach in \cite{10644753} (Phase 3 in Algorithm \ref{alg:hybrid-planning}). This representation ensures continuous derivatives, bounded curvature, and compatibility with quadrotor dynamics while enabling efficient optimization.

\subsubsection{Bezier Curve Representation and Properties} The horizontal trajectory is represented as an $n^{\text{th}}$-degree Bézier curve:
\begin{equation}
\mathbf{r}(s) = \sum_{i=0}^{n} B_{i,n}(s)\,\mathbf{P}_i, \quad s \in [0,1],
\end{equation}
where $\mathbf{P}_0$ and $\mathbf{P}_n$ are fixed start and goal positions, and interior control points $\mathbf{P}_1,\dots,\mathbf{P}_{n-1}$ are optimization variables. The Bernstein basis $B_{i,n}(s) = \binom{n}{i}s^i(1-s)^{n-i}$ ensures the curve lies within the convex hull of its control points—a property exploited for collision avoidance constraints \cite{10644753}.


\subsubsection{Time Parameterization and Cost Functional}

The curve parameter $s$ is mapped to physical time via $t = sT$, where $T$ is the segment traversal time. Time derivatives scale accordingly:
\begin{equation}
\mathbf{v}(t) = \frac{1}{T}\frac{d\mathbf{r}}{ds},\quad 
\mathbf{a}(t) = \frac{1}{T^2}\frac{d^2\mathbf{r}}{ds^2},\quad
\text{snap}(t) = \frac{1}{T^4}\frac{d^4\mathbf{r}}{ds^4}.
\end{equation}

Following \cite{10644753}, the interior control points are optimized to minimize a composite cost:
\begin{equation} \label{eq:bezier_optim}
J = \lambda_p J_{\text{path}} + \lambda_s J_{\text{snap}} + \lambda_t J_{\text{thrust}} + \lambda_w J_{\text{wall}},
\end{equation}
where $J_{\text{path}}$ penalizes deviation from the A* path, $J_{\text{snap}} = \sum \|\text{snap}(t)\|^2$ ensures dynamic smoothness, $J_{\text{wall}}$ enforces obstacle clearance via convex-hull containment, and critically, $J_{\text{thrust}}$ incorporates wind-aware thrust cost as derived below.

\subsubsection{Wind-Aware Thrust Cost}

As shown in \cite{10644753}, for a linear drag model $\mathbf{f}_d = \mathbf{K}(\mathbf{u}_{\text{wind}} - \mathbf{v})$, the thrust objective can be expressed as a quadratic form in the control points:
\begin{equation}
J_{\text{thrust}} = \sum_j (\mathbf{p}_j + \mathbf{v}_j)^\top \mathbf{Q}_W(T_j)(\mathbf{p}_j + \mathbf{v}_j),
\end{equation}
where $\mathbf{v}_j$ encodes wind information regressed from CFD data. This formulation preserves bi-convexity in control points and time intervals, enabling efficient block-coordinate descent optimization \cite{10644753}. Control points are initialized from A* waypoints and constrained to local bounding boxes to maintain topological similarity while allowing refinement for wind-aware smoothness.

\subsubsection{Low-Level Flight Control}

Attitude stabilization, motor mixing, and thrust allocation are executed onboard the quadrotor firmware. 
The planning framework provides position and velocity setpoints, while motor commands, battery voltage, and state estimates are logged for flight analysis. Our control and tracking pipeline is shown in Figure~\ref{fig:tracking_pipeline}.

\section{Evaluation}\label{sec:evaluation}


We evaluated two algorithms, \ALG{} (ours) and \BASE{}, across three real-world experiments. Both algorithms utilized identical Bezier curve fitting and cost functions to optimize for smoothness, obstacle avoidance, and start-to-goal trajectory. The key difference between the algorithms was \ALG{} generated wind-\textit{aware} trajectories, while \BASE{} generated wind-\textit{agnostic} trajectories, where \BASE{} prioritized planning the shortest path. We intentionally created environments where the shortest path was more direct, but more turbulent via crosswinds or resistant via headwinds. Our objective was to determine if \ALG{}, a fluid-aware approach, would produce more stable, safe, and energy efficient paths compared to \BASE{}, a standard shortest-path planner.


\begin{figure}
    \centering
    \vspace{4pt}
    \includegraphics[width=0.75\linewidth]{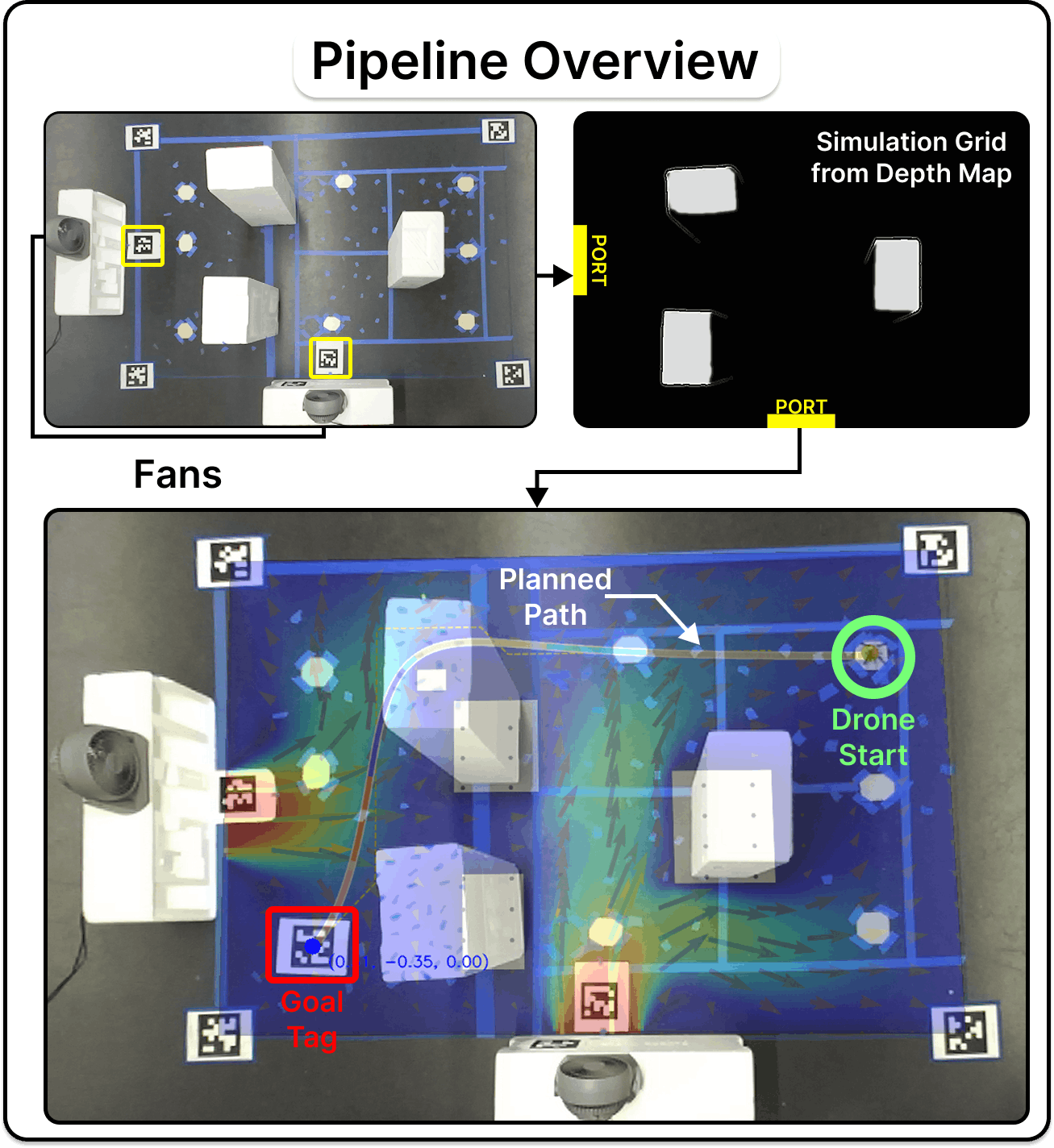}
    \caption{Experimental pipeline. We capture an image of the environment's physical space and the positions of the fans, obstacles, and drone's start and goal (top left). We convert the image to an occupancy grid with fan positions (1s), and use that for our simulation (4s). Finally, we compute the drone's path (5s) and, while the drone flies, localizes it (bottom).}
    \label{fig:experimental_pipeline}
\end{figure}

\subsubsection{Physical Setup} The 3 m x 1.8 m experimental arena consisted of an overhead camera, household fans, and styrofoam obstacles (Figure~\ref{fig:experimental_pipeline}). Flights were performed using a Crazyflie 2.1 UAV equipped with a Flow V2 VIO sensor. To enhance VIO tracking accuracy, we randomly placed colored tape on the floor.

\begin{figure}[H]
    \centering
    \includegraphics[width=0.9\linewidth]{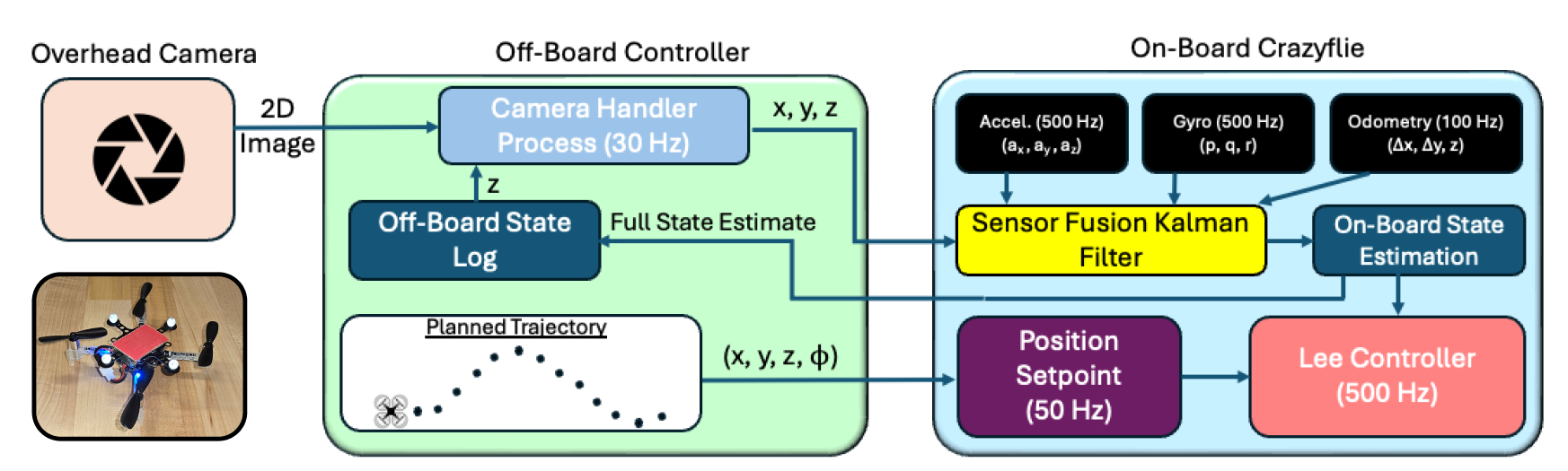}
    \caption{We estimate the state of the Crazyflie drone by fusing odometry
from its onboard Flow V2 sensor and position data from the overhead
camera via HSV tracking.}
    \label{fig:tracking_pipeline}
\end{figure}

We introduced controlled instability using household fans positioned along the environment's perimeter. Table~\ref{tab:fan_speed} shows the wind speed of each fan relative to a distance and power setting, which were measured by an anemometer. Styrofoam boxes (1.0x0.4x0.25 m) served as obstacles to either block or funnel wind and were oriented horizontally and vertically.

The environment data was captured using a ZED Stereo 2i camera (720p at 30 fps). We generated a map via depth filtering to find obstacles and using AprilTags to locate fan (for wind reports) and the start and goal positions. Following perspective and distortion correction, these coordinates were parsed into C++ for simulation in \fluidsim, with flow fields exported as VTK files. During the experiments, we tracked the drone’s position via HSV tracking, fused with VIO sensor readings which was fed into an EKF filter.  (Figure~\ref{fig:tracking_pipeline}). 



\begin{figure}
    \centering
    \vspace{4pt}
    \includegraphics[width=0.88\linewidth]{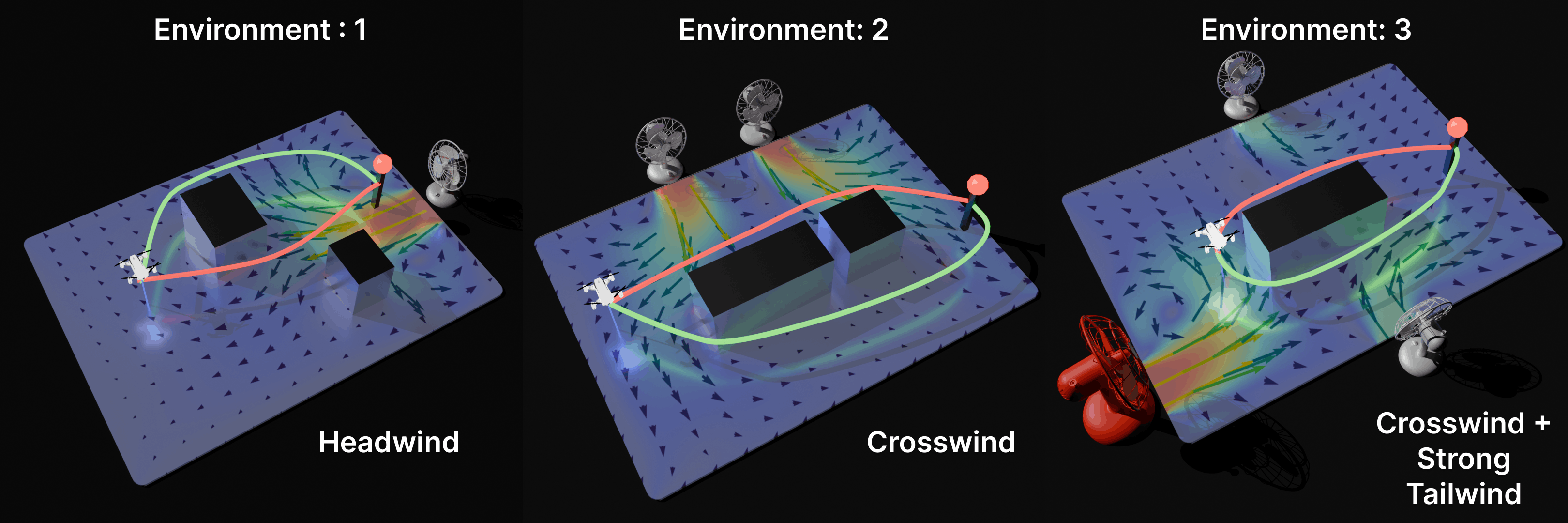}
    \caption{3D illustration of our three environments with the drone at the start and the red circle as the goal. The red fan represents high-speed, while the white fan represents low-speed. The \BASE {} path is shown as a red line, while the \ALG{} path is green. The velocities shown in each environment are relative values, not absolute.}
    \label{fig:environments}
\end{figure}

\begin{table}[t]
    \centering
    \caption{Fan Speed Measurements over Distance}
    \label{tab:fan_speed}
    \begin{tabular}{@{} S[table-format=1.1] S[table-format=1.1] S[table-format=1.1] @{}}
        \toprule
        {\textbf{Distance (m)}} & {\textbf{Low-Speed (m/s)}} & {\textbf{High-Speed (m/s)} } \\
        \midrule
        0.5 & 4.1 & 5.5 \\
        1.0 & 3.2 & 3.8 \\
        1.5 & 1.7 & 2.7 \\
        3.0 & 0.7 & 1.8 \\
        \bottomrule
    \end{tabular}
\end{table}

For our experiments, we created three environments (Figure~\ref{fig:environments}): \ENV[*]{6}: Direct Headwind Avoidance, \ENV[*]{3}: Turbulent Crosswinds, and \ENV[*]{4}: Tailwind-Assisted Obstacle Avoidance.

\section{Results}\label{sec:results}
To establish a performance baseline, we first evaluated both the \BASE{} and \ALG{} trajectories under calm conditions (``No wind'') to measure the inherent tracking error of the Lee controller. We then repeated the trials with the fans on (``Wind''), recording five flights per path. Comparing each path's ``Wind'' results to its ''No Wind'' baseline isolates the effect of disturbances and quantifies the resilience provided by \ALG{} under controlled conditions.

\begin{figure}[t]
    \centering
    \vspace{4pt}
    
    \includegraphics[width=0.9\linewidth]{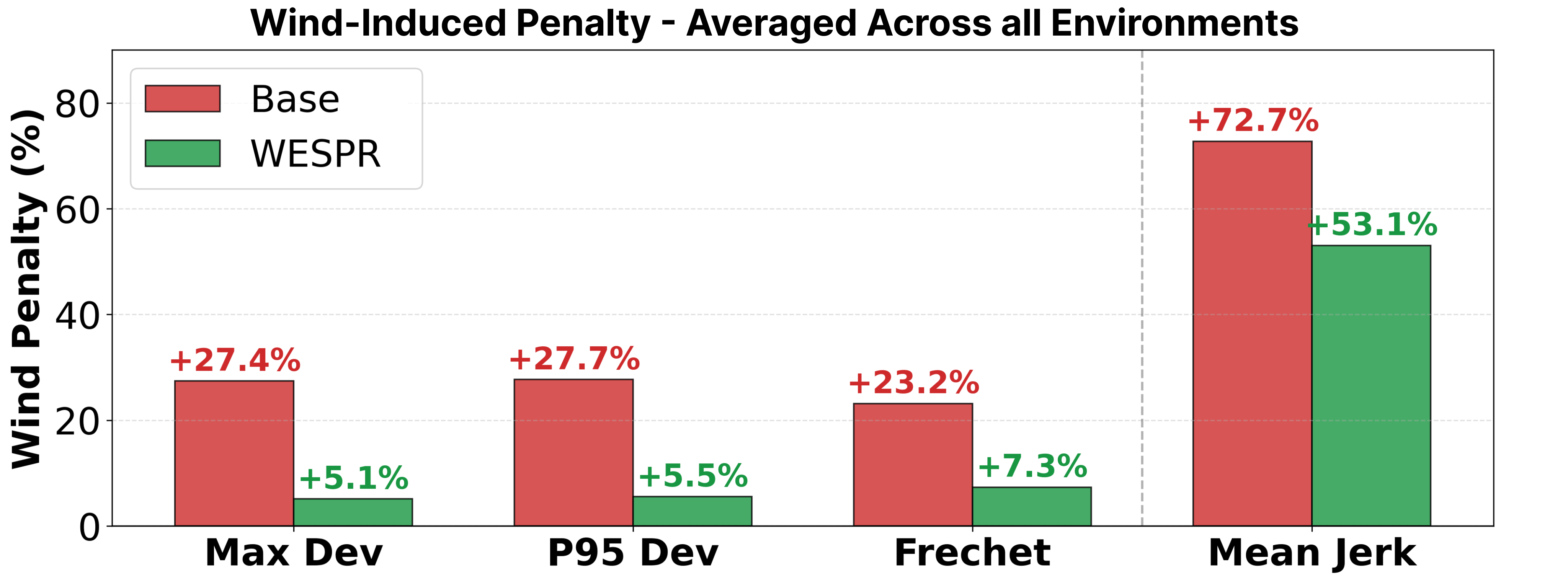}
    
    \caption{Average wind-induced performance metrics across all environments (lower is better).}
    \label{fig:wind_penalty}
\end{figure}


 \subsection{Quantitative Results}\label{sec:results:quantitative}
 

Table \ref{tab:wespr_results} shows how much each metric (maximum deviation, 95th percentile (P95) deviation, and Fréchet distance) worsens with wind for \BASE{} versus \ALG{}, and how much \ALG{} reduces that degradation relative to \BASE{}. Larger positive relative reduction (Rel. Red.) values mean \ALG{} is more robust in that environment and metric, keeping trajectories closer to their no-wind baseline than the \BASE{} planner. The "Val." column represents the absolute Setpoint Tracking Error, measuring the 3D distance between the drone's actual position and its intended reference path. We analyzed maximum and P95 deviations to evaluate susceptibility to catastrophic failure under wind disturbances~\cite{4739221}. To separate temporal lag from structural path divergence, we also computed the discrete Fréchet distance, which quantifies topological similarity between reference and executed trajectories~\cite{alt1995}. Such disturbance-based analyses are standard for assessing controller robustness~\cite{drones6120379}. Under wind, \ALG{} consistently outperformed \BASE{}, achieving up to $58.7\%$ lower maximum deviation, $57.9\%$ lower P95 deviation, and a $46.9\%$ improvement in Fréchet distance (Table~\ref{tab:wespr_results}). Overall, across all trajectories and environments, the aggregrate, wind-induced penalty for all metrics is significantly lower for \ALG{} compared to \BASE{}, as seen in Figure~\ref{fig:wind_penalty}.

\begin{table}[h]
\caption{Wind-Induced Penalty (\%) by Environment (Best Emphasized)}
\label{tab:wind_penalty_summary}
\centering
\begin{tabular}{cccc}
\toprule
\textbf{Env} & \textbf{Metric} & \textbf{Pen.($\mathcal{P}^{\BASE{}}_{wind}$)} & \textbf{Pen.($\mathcal{P}^{\ALG{}}_{wind}$)} \\ \midrule
\multirow{2}{*}{\ENV[*]{6}} & Max Dev  & 51.5 & \textbf{20.3} \\
 & P95 Dev  & 50.8 & \textbf{18.0} \\
 & Frechet  & 37.8 & \textbf{22.1} \\ \midrule
\multirow{2}{*}{\ENV[*]{3}} & Max Dev  & 76.8 & \textbf{42.8} \\
 & P95 Dev  & 67.4 & \textbf{35.1} \\
 & Frechet  & 97.5 & \textbf{50.0} \\ \midrule
\multirow{2}{*}{\ENV[*]{4}} & Max Dev  & \textbf{66.8} & 70.7 \\
 & P95 Dev  & \textbf{60.6} & 60.9 \\
 & Frechet  & 89.9 & \textbf{66.2} \\ \bottomrule
\end{tabular}
\end{table}

\begin{table}[t]
\caption{\ALG{} Quantitative Performance Comparison}
\label{tab:wespr_results}
\centering
\setlength{\tabcolsep}{1pt}

\begin{tabular}{ccccccc}

\toprule
\textbf{Env} & \textbf{Metric} & \textbf{Case} & \textbf{Mode} & \textbf{Val.(m)} & \textbf{Traj. Disp.($\Delta$;m)} & \textbf{Rel.Red.($\eta_r$\%)} \\ \midrule
\multirow{12}{*}{\ENV[*]{6}} & \multirow{4}{*}{\shortstack[c]{Max\\Dev}} & \multirow{2}{*}{\BASE{}} & No Wind & 0.386 & \multirow{2}{*}{0.199} & \multirow{4}{*}{\textbf{58.7\%}} \\ 
                     &                          &                           & Wind    & 0.483 &                        &                                  \\ \cmidrule{3-6}
                     &                          & \multirow{2}{*}{\ALG{}}    & No Wind & 0.405 & \multirow{2}{*}{0.082} &                                  \\ 
                     &                          &                           & Wind    & 0.430 &                        &                                  \\ \cmidrule{2-7}
                     & \multirow{4}{*}{\shortstack[c]{P95\\Dev}} & \multirow{2}{*}{\BASE{}} & No Wind & 0.336 & \multirow{2}{*}{0.171} & \multirow{4}{*}{\textbf{57.9\%}} \\ 
                     &                          &                           & Wind    & 0.440 &                        &                                  \\ \cmidrule{3-6}
                     &                          & \multirow{2}{*}{\ALG{}}    & No Wind & 0.399 & \multirow{2}{*}{0.072} &                                  \\ 
                     &                          &                           & Wind    & 0.417 &                        &                                  \\ \cmidrule{2-7}
                     & \multirow{4}{*}{Frechet} & \multirow{2}{*}{\BASE{}} & No Wind & 0.337 & \multirow{2}{*}{0.128} & \multirow{4}{*}{\textbf{46.9\%}} \\ 
                     &                          &                           & Wind    & 0.440 &                        &                                  \\ \cmidrule{3-6}
                     &                          & \multirow{2}{*}{\ALG{}}    & No Wind & 0.307 & \multirow{2}{*}{0.068} &                                 \\ 
                     &                          &                           & Wind    & 0.296 &                        &                                 \\ \midrule
\multirow{12}{*}{\ENV[*]{3}} & \multirow{4}{*}{\shortstack[c]{Max\\Dev}} & \multirow{2}{*}{\BASE{}*} & No Wind & 0.297 & \multirow{2}{*}{0.228} & \multirow{4}{*}{\textbf{44.3\%}} \\ 
                     &                          &                           & Wind    & 0.409 &                        &                                  \\ \cmidrule{3-6}
                     &                          & \multirow{2}{*}{\ALG{}}    & No Wind & 0.297 & \multirow{2}{*}{0.127} &                                  \\ 
                     &                          &                           & Wind    & 0.314 &                        &                                  \\ \cmidrule{2-7}
                     & \multirow{4}{*}{\shortstack[c]{P95\\Dev}} & \multirow{2}{*}{\BASE*{}} & No Wind & 0.293 & \multirow{2}{*}{0.197} & \multirow{4}{*}{\textbf{49.5\%}} \\ 
                     &                          &                           & Wind    & 0.386 &                        &                                  \\ \cmidrule{3-6}
                     &                          & \multirow{2}{*}{\ALG{}}    & No Wind & 0.284 & \multirow{2}{*}{0.100} &                                  \\ 
                     &                          &                           & Wind    & 0.303 &                        &                                  \\ \cmidrule{2-7}
                     & \multirow{4}{*}{Frechet} & \multirow{2}{*}{\BASE*{}} & No Wind & 0.232 & \multirow{2}{*}{0.226} & \multirow{4}{*}{\textbf{43.8\%}} \\ 
                     &                          &                           & Wind    & 0.211 &                        &                                  \\ \cmidrule{3-6}
                     &                          & \multirow{2}{*}{\ALG{}}    & No Wind & 0.254 & \multirow{2}{*}{0.127} &                                 \\ 
                     &                          &                           & Wind    & 0.301 &                        &                                 \\ \midrule
\multirow{12}{*}{\ENV[*]{4}} & \multirow{4}{*}{\shortstack[c]{Max\\Dev}} & \multirow{2}{*}{\BASE{}*} & No Wind & 0.389 & \multirow{2}{*}{0.260} & \multirow{4}{*}{\textbf{12.5\%}} \\ 
                     &                          &                           & Wind    & 0.464 &                        &                                  \\ \cmidrule{3-6}
                     &                          & \multirow{2}{*}{\ALG{}}    & No Wind & 0.322 & \multirow{2}{*}{0.228} &                                  \\ 
                     &                          &                           & Wind    & 0.332 &                        &                                  \\ \cmidrule{2-7}
                     & \multirow{4}{*}{\shortstack[c]{P95\\Dev}} & \multirow{2}{*}{\BASE{}*} & No Wind & 0.367 & \multirow{2}{*}{0.222} & \multirow{4}{*}{\textbf{14.8\%}} \\ 
                     &                          &                           & Wind    & 0.442 &                        &                                  \\ \cmidrule{3-6}
                     &                          & \multirow{2}{*}{\ALG{}}    & No Wind & 0.311 & \multirow{2}{*}{0.190} &                                  \\ 
                     &                          &                           & Wind    & 0.327 &                        &                                  \\ \cmidrule{2-7}
                     & \multirow{4}{*}{Frechet} & \multirow{2}{*}{\BASE{}*} & No Wind & 0.278 & \multirow{2}{*}{0.250} & \multirow{4}{*}{\textbf{36.9\%}} \\ 
                     &                          &                           & Wind    & 0.412 &                        &                                  \\ \cmidrule{3-6}
                     &                          & \multirow{2}{*}{\ALG{}}    & No Wind & 0.238 & \multirow{2}{*}{0.158} &                                 \\ 
                     &                          &                           & Wind    & 0.254 &                        &                                 \\ \bottomrule
\end{tabular}
\end{table}



\noindent \textbf{Wind Robustness Metrics.} We quantify robustness via \textit{Trajectory Displacement} ($\Delta$), the maximum geometric shift between wind ($\mathbf{p}_{\text{wind}}$) and reference ($\mathbf{p}_{\text{no\_wind}}$) trajectories, for each planner $p \in \{\text{\BASE{}, \ALG}\}$:
\begin{equation}
    \Delta_p = \max_{t} \left\| \mathbf{p}_{\text{wind}}^{p}(t) - \mathbf{p}_{\text{no\_wind}}^{p}(t) \right\|_2
\end{equation}
The relative reduction ($\eta_r$) of \text{\ALG} over \text{\BASE} is defined as:
\begin{equation}
    \eta_r = \frac{\Delta_{\text{\BASE}} - \Delta_{\text{\ALG}}}{\Delta_{\text{\BASE}}} \times 100\%
\end{equation}
Second, we calculate the Wind-Induced Penalty ($\mathcal{P}$) for any scalar metric $M$ (e.g., Max Deviation, Mean Jerk) to measure the relative performance degradation:
\begin{equation}
\mathcal{P}_{\text{wind}}^{p}(M) =
\frac{M_{\text{wind}}^{p} - M_{\text{no\_wind}}^{p}}{M_{\text{no\_wind}}^{p}}
\times 100\%
\end{equation}

\begin{figure}[t]
    \centering
    \vspace{4pt}
    \includegraphics[width=0.88\linewidth]{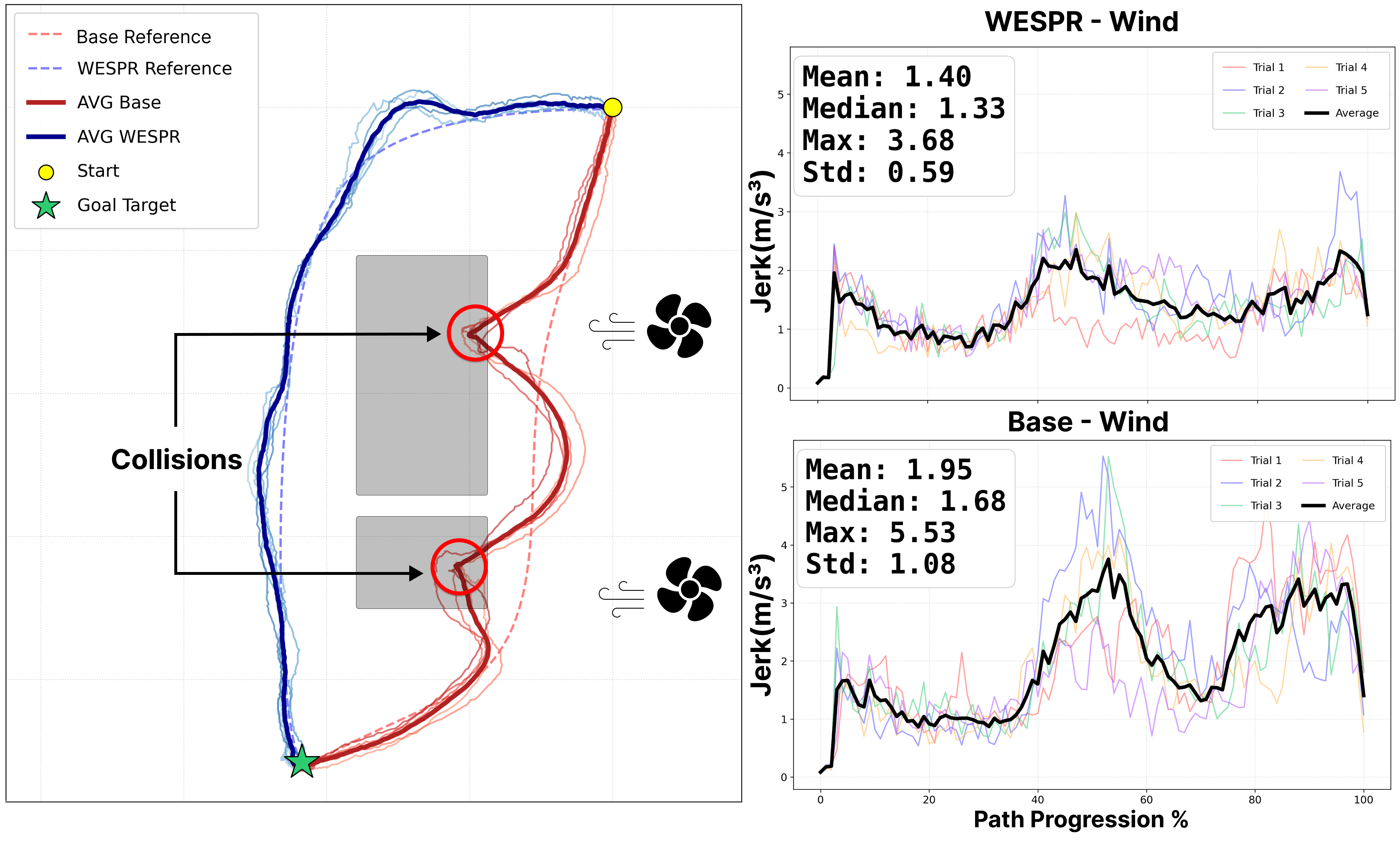}
    \caption{(LEFT) Actual trajectories taken in \ENV[*]{3} by the drone (faintly visible with average path in bold), \ALG{} (Blue) and \BASE{} (Red). \BASE{} trajectory collides with the obstacles. (RIGHT) Jerk Magnitude along the path. \ALG{} path is more stable with reduced max and mean jerk compared to \BASE{}.}
    \label{fig:env2_wind_analysis}
\end{figure}

A key note regarding trajectories marked as \BASE{}* in Table~\ref{tab:wespr_results} is that these trials were conducted in a \emph{simplified environment}, where we removed obstacles in \ENV[*]{3} and \ENV[*]{4} or deactivated certain fans in \ENV[*]{4}. This adjustment prevented 

repeated collisions caused by crosswind fans along the trajectory that we encountered in our experimental trials, allowing us to still obtain quantitative data on the overall path taken and trajectory deviation. Figure~\ref{fig:env2_wind_analysis} shows that the \BASE{} path was pushed into obstacle spaces at two points, where crosswinds induced collisions (this also happened in \ENV[*]{4}). Because a tailwind and crosswind fan on the opposite side were normally blocked by the obstacle, both were turned off when it was removed.


The consequences of neglecting aerodynamic effects are clear in the trajectory analysis of \ENV[*]{3} (Figure~\ref{fig:env2_wind_analysis}). The \BASE{} trajectory failed under concentrated crosswinds between nearby obstacles, deviating sharply and resulting in repeated collisions. This behavior aligned with the degradation reported in Table~\ref{tab:wind_penalty_summary}, where \BASE{} experiences wind-induced penalties of $76.8\%$ for max deviation and $67.4\%$ for the $95^{\text{th}}$ Percentile Error. Because \BASE{} reacted after displacement, actuator saturation produced two distinct jerk peaks twoards the middle and end of the flight (Figure~\ref{fig:env2_wind_analysis} right). In contrast, \ALG{} anticipated disturbances through wind-aware optimization, completing the trajectory without collisions and achieving lower wind-induced penalties in Maximum Deviation ($42.8\%$) and Fréchet Distance ($50.0\%$).

To evaluate actuator effort and overall flight smoothness, we analyzed the Mean Jerk across all trajectories (Table~\ref{tab:merged_results}). \ALG{} significantly reduced the control effort required to stabilize the drone under wind, with jerk reductions of $22.6\%$ for \ENV[*]{6} and $24.6\%$ for \ENV[*]{3}. This effect is especially visible for the \BASE{} path in \ENV[*]{3}, where the spatial jerk profile (Figure~\ref{fig:env2_wind_analysis}) reveals severe wind-induced perturbations as it blindly battles a concentrated crosswind. By anticipating these turbulent regions, the \ALG{} planner prevents the onboard controller from executing the harsh reactive corrections typical of standard tracking in adverse flows.


However, \ALG{} exhibited a $25.5\%$ increase in jerk under wind in \ENV[*]{4} as it seeks local tailwinds to maximize efficiency. Although \BASE{}* had lower jerk, this trajectory was recorded without obstacles and would have resulted in collisions if they were present. In contrast, the \ALG{} trajectory was jerkier but collision-free. We also observed that disabling the tailwind fan immediately caused collisions, whereas no collisions occurred when it was on, indicating that the planner correctly anticipated and leveraged tailwinds to enhance overall UAV safety.

\begin{table}[t]
\centering
\caption{Comprehensive Performance and Dynamics Analysis}
\label{tab:merged_results}
\setlength{\tabcolsep}{4pt}
\footnotesize
\begin{tabular}{@{}llcccr@{}}
\toprule
\textbf{Env} & \textbf{Metric} & \textbf{\BASE{}} & \textbf{\ALG{}} & \textbf{Rel./Reduc.(\%)} \\
\midrule

\multirow{3}{*}{\ENV[*]{6}}
& Path Length (m) & 2.785 & 3.211 & +15.29\% \\
& Mean Jerk ($m/s^3$) & 1.94 & 1.50 & +22.64\% \\
& Collision & No & No & -- \\
\midrule

\multirow{3}{*}{\ENV[*]{3}*}
& Path Length (m) & 3.326 & 3.484 & +4.75\% \\
& Mean Jerk ($m/s^3$) & 1.63 & 1.23 & +24.62\% \\
& Collision & Yes & No & -- \\
\midrule

\multirow{3}{*}{\ENV[*]{4}*}
& Path Length (m) & 2.878 & 3.170 & +10.15\% \\
& Mean Jerk ($m/s^3$) & 1.05 & 1.32 & -25.51\% \\
& Collision & Yes & No & -- \\
\bottomrule
\end{tabular}
\end{table}


\subsection{Qualitative Results}\label{sec:results:qualitative}

Beyond quantitative tracking accuracy, we also evaluated the end-to-end computational feasibility of the proposed pipeline. The complete \ALG{} pipeline executes in as little as \SI{10}{s} on an NVIDIA RTX 3080, with approximately one second for environment capture, four seconds for fluid simulation, and five seconds for trajectory planning. On a lower-tier GPU (NVIDIA GTX 1650), the total runtime remains within \SI{15}{s} (one second for capture, seven seconds for simulation, seven seconds for planning). These results indicate that \ALG{} can be integrated into existing systems with minimal computational overhead while providing meaningful improvements in safety, stability, and energy usage.




Finally, we assessed mission-level robustness in terms of collision-free completion (Table~\ref{tab:merged_results}). While the \BASE{} planner successfully handled simple headwind scenarios, it failed in \ENV[*]{3} and \ENV[*]{4}, where collisions occurred due to stronger wind disturbances along the path. In contrast, \ALG{} reached the goal in all tested environments, underscoring its effectiveness as a safety-critical planning layer for UAV operation in complex wind fields.



\subsection{Ablations}
\subsubsection{CFD Simulation Comparison}

\begin{figure}[t]
    \centering
    \vspace{4pt}
    \includegraphics[width=0.75\linewidth]{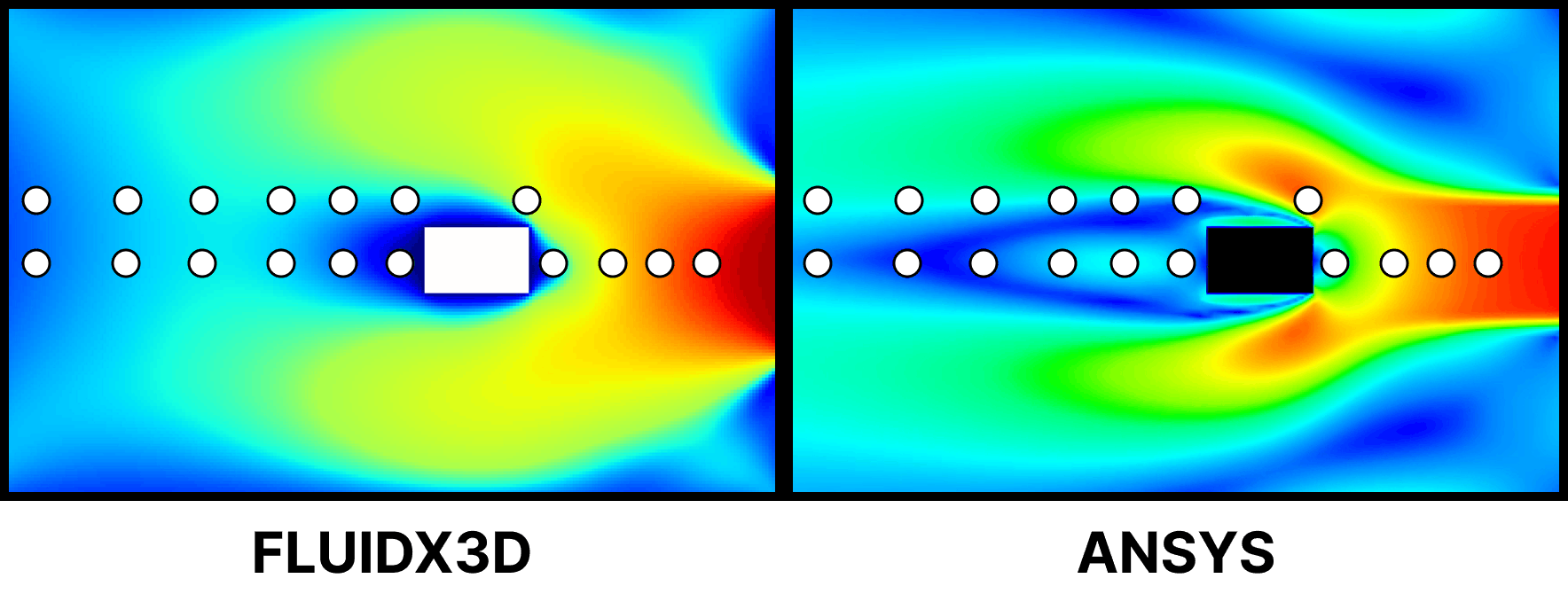}
    \caption{Comparison of FluidX3D and Ansys Fluent configured with matching boundary conditions ($Re$, $V_{\text{in}}$), across 17 probe locations (white circles) against real-world anemometer measurements.} 
    \label{fig:ansys}
\end{figure}


To validate the simulation, we modeled flow around an obstacle using both Ansys Fluent and our pipeline, and compared the results with anemometer measurements in the test environment. Ansys achieved higher accuracy with an average error of $\pm 0.3$~m/s across 17 zones, while FluidX3D (scaled by fan speed) produced $\pm 0.5$~m/s. Despite this, the simulated flow structure was similar (Fig.~\ref{fig:ansys}), capturing key effects such as upstream velocity reduction and flow convergence/divergence around the obstacle. The main discrepancy was an overestimation of velocity magnitudes in \fluidsim, which effectively introduces a conservative safety margin for planning. Some error may also arise from the linear scaling based on maximum fan speed. Given the significantly shorter runtime of our pipeline compared to the more involved Ansys workflow (manual geometry setup, meshing, and $\sim$2 min solve time), this accuracy is sufficient for fast wind-aware planning.

\subsubsection{Bezier vs A* Path Stability}

\begin{figure}[t]
    \centering
    \vspace{4pt}
    \includegraphics[width=0.75\linewidth]{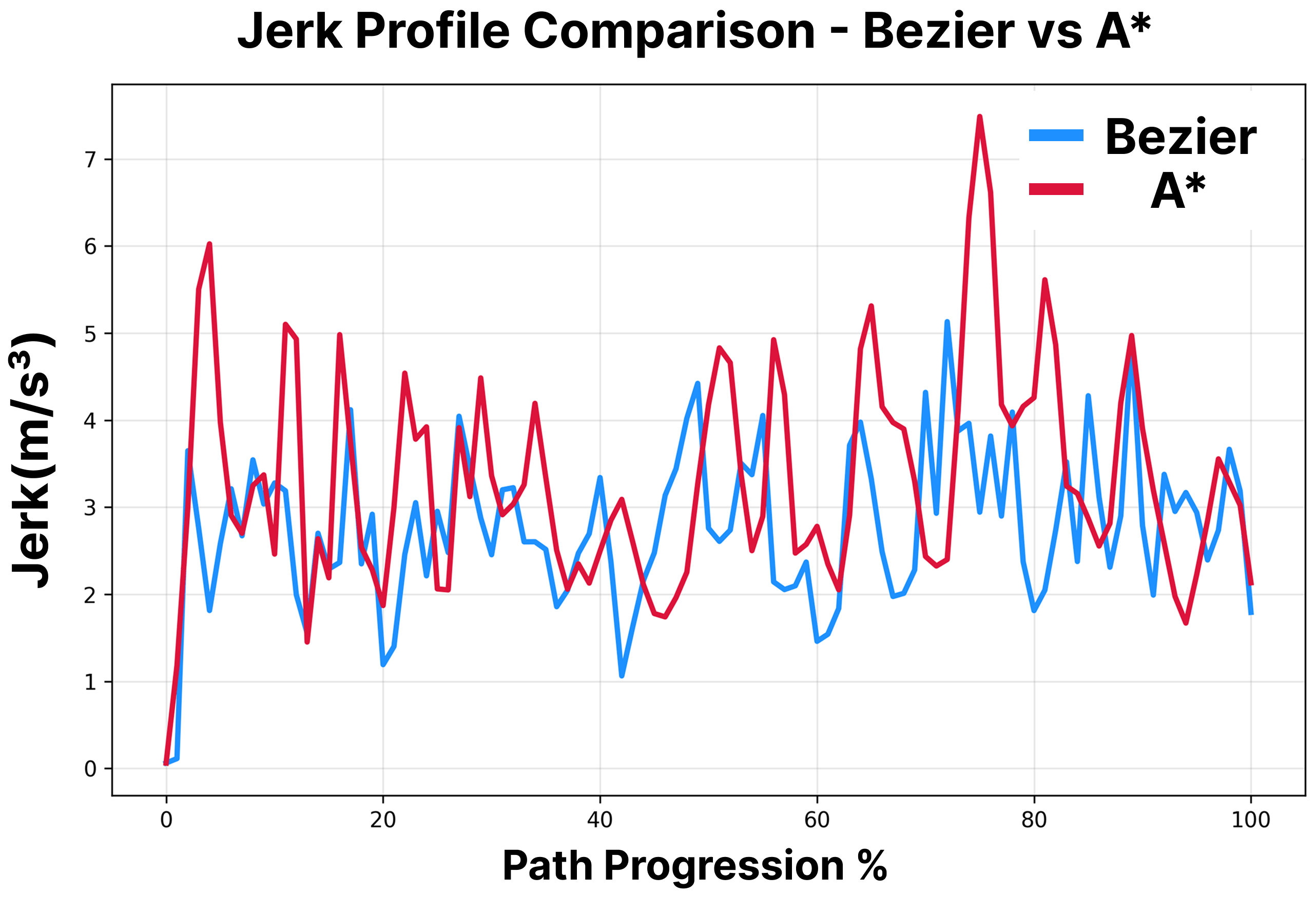}
    \caption{A Comparison of Jerk between Bezier vs. A* Trajectory.}
    \label{fig:Jerk_A*}
\end{figure}


We verified that Bézier smoothing has a meaningful effect on the jerk experienced by the drone (Figure~\ref{fig:Jerk_A*}), with nearly a 20\% reduction indicating an overall smoother flight and improved tracking, especially at higher velocities. We also observed a 36\% gain in reference path tracking when using the Bézier curve.

\subsection{Discussion}

The Crazyflie platform doesn't provide power usage metrics, so we use jerk as a proxy for control effort and actuator strain, similar to \cite{rapakoulias2024stochastic}, which uses RMS angular acceleration. When the vehicle blindly fights severe crosswinds, low-level controllers violently modulate motor RPMs, generating the severe jerk spikes seen in the \BASE{} profiles (Fig. 7). WESPR resolves this with paths $15.29\%$ and $10.15\%$ longer in \texttt{Environment-1} and \texttt{Environment-3}, but inherently smoother; routing through sheltered, low-velocity wake regions eliminates the sustained peak thrust needed to battle headwinds along shorter paths, reducing actuator strain while guaranteeing flight safety. We also attempted to integrate Contextual NeuroMHE \cite{Torshizi2025ContextualNeuroMHE}, but the small testing area and rapidly shifting conditions exceeded its observation window, and the resulting estimation latency caused a jerkier flight.\textbf{Limitations --} Our pipeline has two limitations. First, cost function (\ref{sec:method:costmapgen}) and Bézier optimization parameters were tuned experimentally per environment, using qualitative trajectory analysis and thrust efficiency metrics. Second, we simulate a single 2D plane rather than a full 3D wind field; this was reasonable since our obstacles were uniform in height, but a general UAV deployment would require 3D geometry and wind sources.


\section{Conclusion}\label{sec:conclusion}



In this paper, we presented \ALG{}, a CFD-aware trajectory planning pipeline that integrates fast Lattice Boltzmann simulations via \fluidsim~into quadrotor path optimization. By embedding wind field priors directly into the cost function, \ALG{} generates trajectories with reduced deviation, lower jerk, improved obstacle clearance, and enhanced robustness to adverse winds compared to the geometry-only baseline \BASE{}. Our results show that lightweight, high-speed CFD solvers can provide useful aerodynamic priors for safer, more stable trajectory planning. In particular, \fluidsim's efficiency enables near real-time flow estimation, making wind-aware optimization feasible even on modest hardware. Our validation, however, is limited to a controlled indoor setting with simplified geometry, which, while offering a repeatable proxy for wind–structure interactions, does not capture the full complexity of outdoor environments; convergence and steady-state behavior may differ in larger domains with richer boundary conditions. Nonetheless, the solver’s speed suggests that localized flow simulations can be applied incrementally over regions of interest, supporting scalable deployment.

\addtolength{\textheight}{0cm}   





\bibliographystyle{IEEEtran}
\bibliography{root}

\end{document}